# A Technical Report for VIPriors Image Classification Challenge

Zhipeng Luo, Ge Li, Zhiguang Zhang
DeepBlue Technology (Shanghai) Co., Ltd
{luozp, lige, zhangzhg}@deepblueai.com

*Abstract*—Image classification has always been a hot and challenging task. This paper is a brief report to our submission to the VIPriors Image Classification Challenge. In this challenge, the difficulty is how to train the model from scratch without any pretrained weight. In our method, several strong backbones and multiple loss functions are used to learn more representative features. To improve the models' generalization and robustness, efficient image augmentation strategies are utilized, like autoaugment and cutmix. Finally, ensemble learning is used to increase the performance of the models. The final Top-1 accuracy of our team DeepBlueAI is 0.7015, ranking second in the leaderboard.

## I. INTRODUCTION

The VIPriors Image Classification Challenge is one of "Visual Inductive Priors for Data-Efficient Computer Vision" challenges. The main objective of the challenge is to obtain the highest Top-1 Accuracy on Imagenet dataset. The training and validation data are two subsets of the training split of the Imagenet 2012. The test set is taken from the validation split of the Imagenet 2012 dataset. Each data set includes 50 images per class.

In recent years, a lot of work has achieved excellent results on the Imagenet dataset. The seminal ResNet models [1], introduced in 2016, revolutionized the world of deep learning. ResNeXt [2] adopts group convolution [3] in the ResNet bottle block, which converts the multi-path structure into a unified operation. SE-Net [4] introduces a channel-attention mechanism by adaptively recalibrating the channel feature responses. ResNeSt [5] introduces a Split-Attention block that enables attention across feature-map groups. TResNet [6] introduces a series of architecture modifications that aim to boost neural networks' accuracy while retaining their GPU training and inference efficiency. The above work has inspired us a lot in this Challenge.

Data augmentation is an essential technique for improving the generalization ability of deep learning models. Several strategies have been proposed to automatically search for augmentation policies from a dataset and have significantly enhanced performances on many image recognition tasks, like AutoAugment [7], RandAugment [8].

Compared with the Imagenet dataset of more than 14 million images, the number of images for this challenge is much less. In our method, based on strong backbones, we use multiple loss functions, data augmentation strategies, and ensemble learning to improve classification performance.

## II. METHOD

In this section, our method for this challenge is introduced in detail. The critical parts of our method include model architecture and loss function, which are elaborated in detail as follows.

### A. Model Architecture

We use ResNest101, TResNet-XL and SEResNeXt101 as our backbones. These models are introduced as follows.

**ResNest**: The key part of ResNest is Split-Attention block. Split-Attention block is a computational unit consisting feature-map group and split attention operations. Figure 1 depicts an overview of a Split-Attention Block.

**TResNet**: TResNet design is based on the classical ResNet50 architecture, with dedicated refinements, modifications and optimizations. The refinements mainly include SpaceToDepth

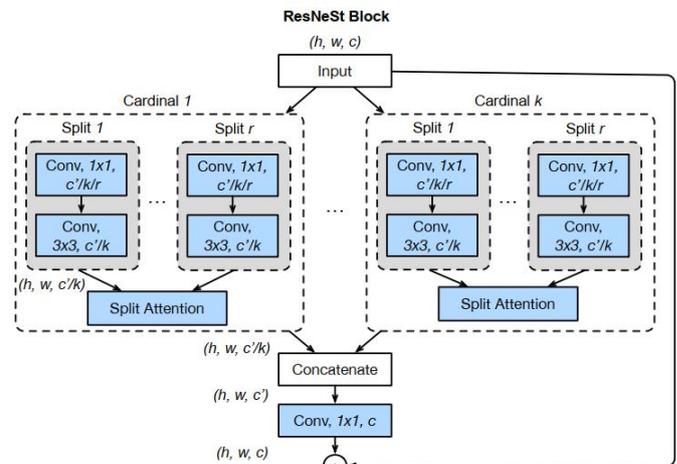

Figure 1 Split-Attention Block of ResNest.

Stem, Anti-Alias Downsampling, In-Place Activated BatchNorm, Block-Type Selection and Optimized SE Layers. The biggest advantage of the network is its high GPU throughput, which enables to practically double the maximal possible batch size. Figure 2 shows TResNet BasicBlock and Bottleneck design.

**SEResNeXt**: ResNeXt is constructed by repeating a building block that aggregates a set of transformations with the same topology. SE block adaptively recalibrates channel-wise feature responses by explicitly modelling interdependencies between channels. SEResNeXt has good performance in multiple tasksand is widely used.

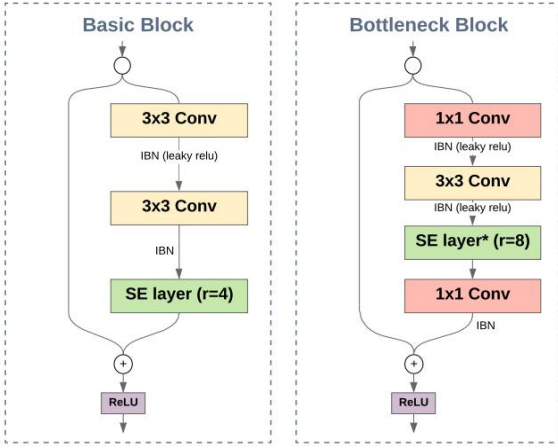

Figure 2 TResNet BasicBlock and Bottleneck design.

*B. Loss function*

**Label Smoothing**: Label smoothing [9] is a mechanism of regularize the classifier layer by estimating the marginalized effect of label-dropout during training. It changes the construction of the true probability to

$$q_i = \begin{cases} 1-\varepsilon & if \ i = y, \\ \varepsilon/(K-1) & otherwise, \end{cases}$$

With label smoothing the distribution centers at the theoretical value and has fewer extreme values.

**Triplet Loss**: Hermans et al. [10] proposed the batch-hard triplet loss that selects only the most difficult positive and negative samples. The formula is as follows:

$$L_1 = \left[ m + \max_{x_p \in P(a)} d(x_a, x_p) - \min_{x_n \in N(a)} d(x_a, x_n) \right]_+$$

**ArcFace loss**: ArcFace loss is proposed to obtain highly discriminative features for face recognition [11]. The calculation is as follows:

$$L_2 = -\frac{1}{N} \sum_{i=1}^{N} \log \frac{e^{s(\cos(\theta_{y_i}+m))}}{e^{s(\cos(\theta_{y_i}+m))} + \sum_{j=1, j \neq y_i}^{n} e^{s \cos \theta_j}}$$

In order to make the network learn better feature representation, in addition to CE loss with label smoothing, batch-hard triplet loss and Arcface loss are used.

In the experiment, we used three types of combined losses, e.g. CE loss with label smoothing, CE loss with label smoothing+batch-hard triplet loss, and CE loss with label smoothing+ArcFace loss.

## III. EXPERIMENT

*A. Data augmentation*

Data augmentation can effectively prevent overfitting. The data augmentation strategies we used is described as follows.

**Auto Augmentation:** Auto-Augment is a strategy that augments the training data with transformed images, where the transformations are learned adaptively. A search which tries various candidate augmentation policies returns the best 24 best combinations. One of these 24 policies is then randomly chosen and applied to each sample image during training.

**Cutmix:** Cutmix is a data augmentation strategy that generates a weighted combinations of random image pairs from the training data [12]. Patches are cut and pasted among training images where the ground truth labels are also mixed proportionally to the area of the patches. The Original images and the image after cutmix are shown as Figure 3.

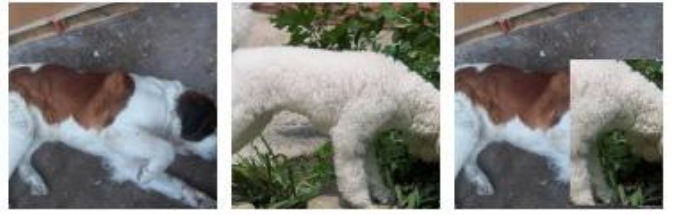

Figure 3. The Original images and the image after cutmix.

During training, we perform the following steps one-by-one:
1. Randomly crop a rectangular region whose aspect ratio is randomly sampled in [3/4, 4/3] and area randomly sampled in [8%, 100%], then resize the cropped region into a square image.
2. Flip horizontally with 0.5 probability.
3. Autoaugment: Randomly choose one of the best 24 Sub-policies on ImageNet.
4. Normalize RGB channels by subtracting the mean value and dividing by the standard deviation.
5. Cutmix with 0.5 probability.

*B. Training strategy*

For the final submission of the competition, 7 models are used, which are ResNest101+CE, ResNest101+CE+triplet loss, ResNest101+CE+Arcface loss, TResNet-XL+CE, TResNet-XL +CE + triplet loss, SEResNeXt101+CE, and SEResNeXt101+CE+Arcface loss.

In the experiment, we used the standard SGD with momentum set to 0.9 in all cases. The batch size is set to 128 for ResNest101 and SEResNeXt101, and 256 for TResNet-XL. Our learning rates are adjusted according to a cosine schedule [13]. The initial learning-rate is set to $\eta = \frac{B}{256}\eta_{base}$, where B is the mini-batch size and we use $\eta_{base} = 0.1$ as the base learning rate. This warm-up [13] strategy is applied over the first epoch, gradually increasing the learning rate linearly from 0 to the initial value for the cosine schedule. The weight decay is set to 0.0001. All the models were implemented in pytorch.

The models with CE loss was trained for 250 epochs. The models with triplet loss or Arcface loss were trained for 200 epochs base on the weights trained using CE loss. This is because we found that it is hard to converge from scratch using CE+triplet loss or CE+Arcface loss.

During testing, the test time augmentation is used. Two cropping methods were used. One is resizing each image's shorter edge to 256 pixels while keeping its aspect ratio and cropping out the 224-by-224 region in the center. The other is randomly cropping a rectangular region whose area is randomly sampled in [80%, 100%]. Each image is input to the networks in four scales, 224, 320, 380, and 448. The cropped image and the horizontally flipped image are input to networks to get results.

*C. Experimental Results*

The training set and the validation set are merged to train our model. And no external images or pre-trained weights are used. The experiment results on test set are shown as Table 1 and Table 2.

Table 1. The experimental results of ResNest101.

| models | loss | train size | test size | Top1-accuracy |
|---|---|---|---|---|
| ResNest101 | CE | 224 | 224 | 0.6316 |
| | | 224 | 320 | 0.6518 |
| | | 224 | 380 | 0.65 |
| | | 448 | 448 | 0.6414 |
| ResNest101 | CE+triplet loss | 224 | 224 | 0.631 |
| | | 224 | 320 | 0.6521 |
| | | 224 | 380 | 0.6476 |
| | | 224 | 448 | 0.6393 |
| ResNest101 | CE+ArcFace loss | 224 | 224 | 0.6335 |
| | | 224 | 320 | 0.6636 |
| | | 224 | 380 | 0.6623 |
| | | 224 | 448 | 0.6583 |

Table 2. The experimental results of TResNet-X and SEResNeXt101.

| models | loss | test size | Top1-accuracy |
|---|---|---|---|
| TResNet- XL | CE | 224 | 0.6058 |
| | | 320 | 0.6368 |
| | | 380 | 0.6345 |
| | | 448 | 0.6418 |
| TResNet- XL | CE+triplet loss | 224 | 0.6262 |
| | | 320 | 0.6516 |
| | | 380 | 0.6482 |
| | | 448 | 0.6399 |
| SEResNeXt101 | CE | 224 | 0.6307 |
| | | 320 | 0.6526 |
| | | 380 | 0.6519 |
| | | 448 | 0.6443 |
| SEResNeXt101 | CE+ArcFace loss | 224 | 0.6303 |
| | | 320 | 0.6584 |
| | | 380 | 0.6594 |
| | | 448 | 0.6532 |

The accuracy in the table is the fusion of the results obtained by the images of the four scales and two cropping methods, that is, the average of the probabilities of the eight results.

For ResNest101+CE, the images with size 448 are used for training. The size of input images is 224 for other experiments.

As can be seen from the tables, the results are improved adding the triplet loss or Arcface loss, indicating the effectiveness of these two losses for the classification task.

In the experiment, we found that setting a large crop size range [8%, 100%] during training, testing with multiple scales has a good effect. The network trained with images of size 224, while images of size 320 or 380 perform best when tested. Fusing the results of multiple scales can further improve the results.

*D. Ensemble Learning*

Experimental evidence shows that the ensemble method is usually much more accurate than a single model. In our method, the ensemble method is averaging the output probability of all prediction results, which are from different models or loss functions.

Although triplet loss or Arcface loss perform better than CE alone, we have found that fusing the results of models with different loss functions has a certain improvement. Therefore, we fuse all the results in Table 1 and Table 2 as the final submission. The 0.7015 is the ensemble results for the above models. The final Top-1 accuracy of our team is 0.7015, ranking second in the leaderboard.

IV. CONCLUSION

In our method, three strong classification models were taken as the backbones. The use of multiple loss functions effectively improves the performance of the models. The image augmentation strategies improve the generalization and robustness of the models and prevents overfitting. Finally, multiple testing methods and ensemble learning effectively improves the final score.